\pdfoutput=1

\documentclass[11pt]{article}

\usepackage[]{acl}

\usepackage{times}
\usepackage{latexsym}
\usepackage{graphicx}
\usepackage{multirow}
\usepackage{booktabs}
\usepackage{amssymb,amsfonts,amsmath}
\usepackage{pifont}
\usepackage{makecell}
\usepackage{float}
\usepackage{xcolor}
\usepackage{url}
\usepackage[nomessages]{fp}
\newcommand{\maxnum}{100.00}
\newlength{\maxlen}
\usepackage[caption=false]{subfig}

\newcommand{\databar}[3]{%
  \settowidth{\maxlen}{\maxnum}%
  \addtolength{\maxlen}{\tabcolsep}%
  \FPeval\result{round(3.15*#2/\maxnum:4)}%
  \rlap{\color{#1}\hspace*{-.5\tabcolsep}\rule[-.35\ht\strutbox]{\result\maxlen}{1.2\ht\strutbox}}%
  \makebox[\columnwidth][c]{#3}%
}
\newcommand{\cmark}{\ding{51}}%
\newcommand{\xmark}{\ding{55}}%

\usepackage[T1]{fontenc}

\usepackage[utf8]{inputenc}

\usepackage{microtype}

\title{Word Sense Extension}

\author{
  Lei Yu$^{1}$, Yang Xu$^{1,\,2}$ \\
  $^1$ Department of Computer Science, University of Toronto \\
  $^2$ Cognitive Science Program, University of Toronto \\
  \texttt{\{jadeleiyu,yangxu\}@cs.toronto.edu}
}

\begin{document}
\maketitle
\begin{abstract}
Humans often make creative use of words to express
novel senses. A long-standing effort in natural language processing has
been focusing on word sense disambiguation (WSD), but little has been explored about how the sense inventory of a word may be extended toward novel meanings. We present a paradigm of \textit{word sense extension} (WSE) that
enables words to spawn new senses toward novel context. We develop a framework that simulates novel word sense extension by first partitioning a polysemous word type into two pseudo-tokens that mark its different senses, and then inferring whether the meaning of a pseudo-token can be extended to convey the sense denoted by the token partitioned from the same word type. Our framework combines cognitive
models of chaining with a learning scheme that transforms a language model embedding space to support
various types of word sense extension. We evaluate our framework
against several competitive baselines and show that it is superior in predicting plausible novel senses for over 7,500  English words. Furthermore, we show that our WSE framework  improves performance over a range of transformer-based WSD models in predicting rare word senses with few or zero mentions in the training data.

\end{abstract}
\section{Introduction}
\begin{figure}[]
 \includegraphics[clip,width=\columnwidth]{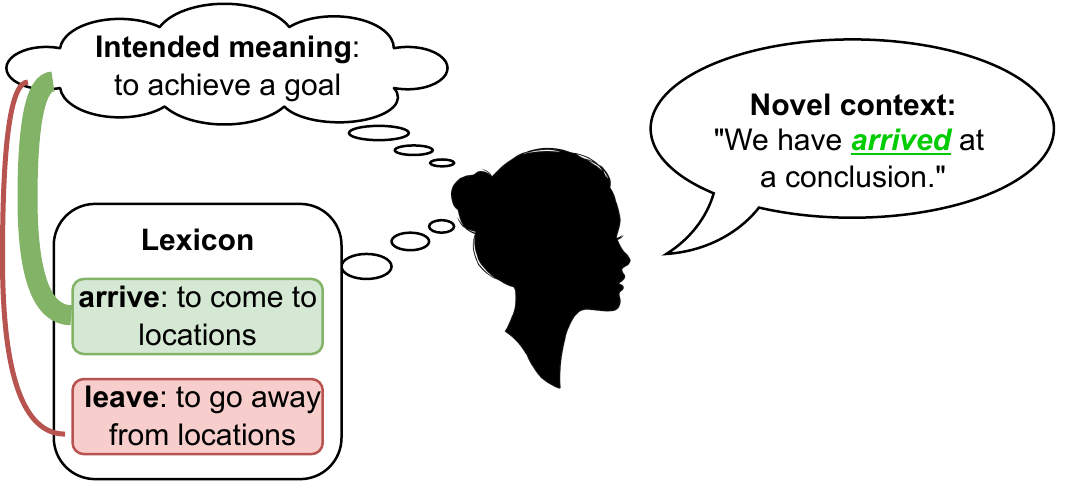}
\caption{Illustration of the problem of word sense extension. Given a novel context, a speaker chooses an existing word in the lexicon to convey a novel intended meaning that has not appeared in the semantics of that word. The speaker determines the appropriateness of a chosen word (indicated by line width of the colored curves) based on semantic relatedness between the novel intended meaning and existing word meanings.}
\label{figure:wse-illustration}
\end{figure}

Humans make creative reuse of words to express novel senses. For example, the English verb {\it arrive} extended from its original sense ``to come to locations (e.g., to \textit{arrive} at the \underline{gate})'' toward new senses  such as ``to come to an event (e.g., to \textit{arrive} at a \underline{concert})'' and ``to achieve a goal or cognitive state (e.g., to \textit{arrive} at a \underline{conclusion})'' (see Figure~\ref{figure:wse-illustration}). The extension of word meaning toward new context may draw on different cognitive processes such as metonymy and metaphor, and here we develop a general framework that infers how words extend to plausible new senses.

A long-standing effort in natural language processing (NLP) is to build systems that support automatic word sense disambiguation (WSD) from linguistic context. This line of work typically takes a discriminative approach toward  word meaning and has developed models relying on both traditional machine learning \cite{gale1992estimating,kilgarriff2000framework,zhong2010makes,iacobacci2016embeddings} and modern neural language models \cite{huang2019glossbert,wiedemann2019does,loureiro2019language,bevilacqua2020breaking}. However, existing WSD models often struggle with recognizing rare word senses with few or no mentions in training \cite{blevins2021fews}. Here we show that by modelling the generative extensional processes of word meaning, WSD models can become better at recognizing infrequent word senses in natural context and without relying on external lexical resources. 

Work in computational and cognitive linguistics shows that word senses  do not extend arbitrarily \cite{nunberg1979non,lehrer1990polysemy,rumshisky-batiukova-2008-polysemy}. Lexical semanticists have suggested that a number of cognitive devices may be applied to generate creative word usages, such as logical metonymy \cite{copestake1995semi,pustejovsky1998generative} and metaphor \cite{lakoff2008metaphors,pustejovsky2010mechanisms}. Cognitive linguists have also suggested that systematic mappings between conceptual domains underlie the metaphorization of word meaning \cite{brugman1988cognitive,lakoff2008metaphors,gentner1983structure}. However, the reliance on hand-crafted rules of semantic  productivity makes it difficult to implement systems that support flexible and scalable extension to new word senses.

We present a paradigm that considers the problem of {\it word sense extension} (WSE) illustrated in Figure \ref{figure:wse-illustration}. Given a novel  context and an intended meaning, a speaker wishes to choose an existing word in the lexicon to express that meaning which the word has never been used to convey. To operationalize a speaker model without prior knowledge about pairings between the novel meaning and existing word forms, we replace each candidate word type with a pair of  ``pseudo-tokens'' that signify one of its existing senses (called the target sense) the other senses (called the source senses) respectively, a method related to previous work in polysemy induction~\cite{pilehvar-navigli-2014-large,dubossarsky-etal-2018-coming}. We then infer whether a partitioned pseudo-token denoting the source sense may be extended to express the target sense denoted by its sibling token partitioned from the same word type. We propose a family of cognitively-inspired probabilistic models for this inference problem. We show that our WSE models can reliably predict plausible novel senses on a large usage-based dataset with approximately 34,000 senses for over 7,500 English word types.\footnote{We release the code and data for our work here: \url{https://github.com/jadeleiyu/word_sense_extension}.}

\section{Related work}

\subsection{Models of word meaning extension}
 Researchers in lexical semantics and cognitive linguistics have both proposed theories to account for the malleable nature of lexical meaning. The Generative Lexicon theory by \citet{pustejovsky1998generative} argues that a fixed set of generative devices, such as type-coercion and co-composition, can operate on the lexical structure a word to produce various related meaning interpretations. \citet{copestake1995semi} also illustrates how formal lexical rules such as grinding and portioning can be applied to produce novel word usages such as logical metonymy. In cognitive linguistics, \citet{lakoff1987women} argues that word meanings grow relying on processes of chaining, whereby novel meanings link to existing ones that are close in semantic space. Similar processes are also relevant to the construction of metaphorical usages in natural language drawing on image schemas \cite{brugman1988cognitive, dewell1994overagain, gibbs2008image} and analogy or structural alignment between domains \cite{gentner1983structure,falkenhainer1989structure}.
 
 Our work builds on the cognitive theory and recent computational work on chaining  \cite{lakoff1987women,malt1999knowing,ramiro2018algorithms,habibi2020chaining,grewal2020chaining,yu2021predicting}, and we show that a chaining-based framework learns systematic patterns of word sense extension discussed in the tradition of generative lexical semantics. Related work has taken a similar approach for modelling sense extension in slang usages \cite{sun2021computational}, but here we consider the more general problem of word sense extension.

\subsection{Models of word sense disambiguation}
A large community in NLP has been working on the problem of word sense disambiguation (WSD). Early WSD systems adopt a knowledge-based approach by comparing the neighborhood context of a target word with its gloss or definition in lexicographic databases such as WordNet \cite{miller1995wordnet, gale1992estimating, kilgarriff2000framework}. Later work develops feature-based classification models to predict sense labels for a word based on its linguistic features \cite{zhong2010makes, iacobacci2016embeddings, raganato2017word}. Recent progress in deep learning also motivates the development of WSD systems based on deep contextualized language models (CLM) or its combination with external lexical knowledge base \cite{huang2019glossbert,hadiwinoto2019improved,bevilacqua2020breaking}. Despite these impressive advances, many CLM-based WSD systems still suffer from the data sparsity that stems from the Zipfian distribution of word senses \cite{kilgarriff2004dominant} -- i.e. the most frequent sense of a polysemous word often accounts for a dominant portion of its mentions, while other senses have much less or even zero frequency in training data. Recent work has proposed to mitigate this sense sparsity problem by resorting to gloss information \cite{luo2018incorporating,kumar2019zero,huang2019glossbert,blevins-zettlemoyer-2020-moving} or non-parametric few-shot learning \cite{holla-etal-2020-learning, chen-etal-2021-non}. We shall demonstrate that learning word sense extensions offers an alternative approach to improve WSD system performance on infrequent word senses by leveraging the systematic semantic relational patterns between conventional and novel word senses.

\subsection{Contextualized semantic representations}
Existing work has proposed to apply contextualized language models to lexical semantic tasks that involve polysemy. Diachronic studies show that contextualized representations of word usage and sense definitions can be used to detect lexical semantic shifts \cite{giulianelli-etal-2020-analysing,hu2019diachronic}. Probing studies also suggest that pretrained contextualized language models encode rich lexical semantic information that may help decide the levels of word polysemy \cite{gari2021let} and infer semantic relations between word senses \cite{vulic-etal-2020-probing}. The WSE paradigm we propose is related to  lexical substitution, where a model is used to replace a target word in a sentence with a substitute word without changing the sentence meaning \cite{mccarthy-navigli-2007-semeval,melamud-etal-2016-context2vec,zhou-etal-2019-bert}. However, our framework goes beyond  this research by asking whether a word can extend its sense inventory to express novel intended meanings in natural context.

\section{Computational framework}
Our framework of word sense extension involves three interrelated components: 1) A procedure for partitioning polysemous words in the lexicon into new pseudo-tokens that signify their different senses; 2) a probabilistic, chaining-based formulation of word sense extension for lexical choice making under novel linguistic context; and 3) a learning algorithm for a transformed semantic space to learn flexible extensions of word senses. 

\subsection{Sense-based word type partitioning} \label{type-partition}
Let $\mathcal{W} = \{w_1,...,w_{|V|}\}$ be our vocabulary of polysemous (English) word types, where each $w$ has a set of $n$ senses $\mathcal{S}_{w} = \{s_{1},...,s_{n}\}$. Assume that for each $w$ there is also a collection of its sense-annotated sample usage contexts $\mathcal{C}_{w} = \{(c_1, y_1),...,(c_m,y_m)\}$, where each contextual sequence $c \in \mathcal{C}_{w}$ is labeled with a sense $y \in \mathcal{S}_{w}$ instantiating the meaning of $w$ in that usage context. We want to simulate the scenario where a speaker, without knowing a priori that a word $w$ has a sense $s^* \in \mathcal{S}_{w}$, is able to extend the meaning of $w$ to expressing $s$ under novel context. 

To operationalize this idea of word sense extension, we first partition each $w$ into two hypothetical tokens: a source token $t^0$ that denotes the set of existing source senses $\mathcal{S}_0 = \mathcal{S} \setminus \{s\}$ of $w$, and a target token $t^*$ that denotes the novel target sense $s^*$ to which $w$ extends beyond its existing senses. We then replace $w$ with $t^0$ in all usage contexts that reflect one of its source senses (i.e., $(c_i, y_i)$ where $y_i \in \mathcal{S}_0$), and replace $w$ with $t^*$ in all usage contexts where $w$ signifies the target sense (i.e. $(c_i, y_i)$ where $y_i = s^*$). 

To guard against information smuggling in predicting novel word sense extension, we learn a contextualized language model from scratch using the set of replaced usage instances. Specifically, the language model is trained on the task of masked language modeling (MLM), where it takes batches of sampled usage instances with some randomly chosen tokens masked out, and updates its parameter weights to maximize the probability of infilling the correct missing tokens. Through this procedure, we obtain a language model that can compute meaningful contextualized representations for the usages of $w$ that instantiate the target sense $s^*$ \textit{without} knowledge that $s$ can be expressed by $w$. 

\subsection{Probabilistic formulation of WSE}
Let $\mathcal{C}_{0}, \mathcal{C}_{*}$ be the two sets of usage instances with $w$ replaced by $t^*$ and $t^0$ respectively. We consider an inference scenario where the language model learned using the procedure from the previous section is presented with a novel usage $c^* \in \mathcal{C}_{*}$ of  target token $t^*$, and is queried to choose among a set of candidate source tokens  to convey the same (and new) intended meaning as that of $t^*$. 

Concretely, suppose the target token $t^*$ partitioned from the verb $w =$ \textit{arrive} denotes its metaphorical sense $s^* = $ ``to achieve a goal'', and the source partitioned token $t^0$ of \textit{arrive} is comprised of its existing source senses (that exclude the metaphorical sense in question). We then use the model to infer whether $t^0$ can be used to convey the new meaning $t^*$ in novel metaphorical usages such as $c= $ ``They finally $t^*$ at a conclusion after a long debate'' (note here the original verb \textit{arrive} is replaced by the target token $t^*$ through word type partitioning). We assess the success of our model by analyzing how it ranks the ground-truth source token (i.e., $t^0$ of {\it arrive}) among the space of alternative candidate source tokens partitioned from other polysemous words in the lexicon. For example, one source token might signify the literal senses of the verb \textit{leave} which differs from the ground-truth verb \textit{arrive}. Formally, we cast WSE as finding a source token $t$ that maximizes the following probability:

\begin{align}
    \text{argmax}_{t} P(t|\textbf{m}(t^*|c^*))
    \label{eq:wse-inference}
\end{align}

Here $\textbf{m}(t^*|c^*)$ is the  representation of target token $t^*$ under context $c^*$ to which $t$ is extended.

\subsection{Chaining-based models of WSE} 

We  present a family of probabilistic  models for Eq.\ref{eq:wse-inference} that draw inspirations from the cognitive theory of chaining~\cite{lakoff1987women,habibi2020chaining}. Our chaining-based WSE models assume that a source token $t^0$ can be extended to express a novel meaning if the new intended meaning is overall similar to $t^0$'s existing senses. We operationalize $\textbf{m}(t^*|c^*)$ as the contextualized word embedding of target token $t^*$ under context $c^*$ computed by the speaker language model, denoted as $\textbf{h}(t^*|c^*)$. We represent the existing senses of source token $t$ as the collection of all of its contextualized embeddings $\textbf{H}(t^0) = \{\textbf{h}(t^0|c)|c \in \mathcal{C}_{0}\}$. The chaining-based WSE models take the general form:
\begin{align}
    P(t^0|\textbf{m}(t^*|c^*)) \propto \text{sim}(\textbf{H}(t^0),\textbf{h}(t^*|c^*))
\end{align}

We consider two common types of chaining model that specify the similarity function $\text{sim}()$.

{\bf WSE-Prototype model.} The prototype model takes inspiration from prototypical network for few-shot learning \cite{snell2017prototypical,holla-etal-2020-learning} and follows the prototype theory of categorization \cite{rosch1975cognitive} in cognitive psychology. It assumes that the existing senses of a source token $t^0$ can be summarized by a global average (i.e., prototype) of its contextualized embeddings in $\textbf{H}(t^0)$, so that the probability of $t^0$ being a good candidate to convey the intended meaning of the target token is proportional to the semantic similarity between the contextualized embedding $\textbf{h}(t^*|c^*)$ of the target token and the prototype of its sibling source token: 

\begin{align}
    &\resizebox{0.85\linewidth}{!}{$P(t^0|\textbf{m}(t^*|c^*)) \propto \exp[-d( \textbf{h}(t^*|c^*),\textbf{z}(t^0))]\label{eq:prototype-sim} $}\\
    &\textbf{z}(t^0)  = \frac{1}{|\mathcal{C}_0|}\sum\limits_{c \in \mathcal{C}_0} \textbf{h}(t^0|c) 
\end{align}  
Here $\textbf{z}(t^0)$ is the global mean contextualized embedding of $t^0$, and we compute dot product as the similarity function $d(\cdot, \cdot)$ between two vectors.\footnote{We experimented with negative squared Euclidean distance suggested in \citet{snell2017prototypical} as an alternative similarity function but found it to yield worse  performance on both WSE and downstream WSD tasks compared to  dot product.} 

{\bf WSE-Exemplar model.} The exemplar model resembles the memory-augmented matching network in deep few-shot learning \cite{vinyals2016matching}, and formalizes the exemplar theory of categorization \cite{nosofsky1986attention}. This model postulates that the meaning of $t^0$ is represented by the collection of its individual usages $c \in \mathcal{C}_{0}$. The probability that $t^0$ can be extended to the meaning $\textbf{m}(t^*|c^*)$ is  proportional to the mean similarity score between $\textbf{h}(t^*|c^*)$ and each contextualized embedding of $t^0$:     
\begin{align}
    \resizebox{\linewidth}{!}{$ P(t^0|\textbf{m}(t^*|c^*)) \propto  \frac{1}{|\mathcal{C}_{0}|}\sum\limits_{c \in \mathcal{C}_{0}} \exp[-d(\mathbf{h}(t^*|c^*),\mathbf{h}(t^0|c))] $} \label{eq:exemplar-sim}
\end{align}

\subsection{Learning sense-extensional semantic space}\label{sense-extension-space}

Chaining relies on identifying close semantic relations between existing senses and generalizing the recognized relations to generate new senses. For instance, if a WSE model has observed how the English verb \textit{grasp} relates its literal sense ``to hold an item firmly" to the extended metaphorical sense ``to understand an idea", the model should also predict similar but novel non-literal sense extensions for other verbs that involve such metaphorical mappings (e.g., the meaning extension of the verb \textit{get} from ``to \textit{get} a car" to ``to \textit{get} someone's idea", which also reflects the conceptual metaphor IDEAS ARE OBJECTS) \cite{lakoff2008metaphors}. 

Following work in deep few-shot learning, we propose an episodic learning algorithm to transform the language model embedding space of the WSE model into a semantic space that better captures the regular, systematic patterns in sense extension. At each episode, we sample a mini-batch of $N$ source-target token pairs $\{(t^0_{i}, t^*_{i})\}_{i=1}^{N}$ partitioned from $N$ distinct polysemous word types, and sample a usage context $c_i^*$ for each target token $t^*_{i}$. The WSE model then chooses the most appropriate source token to convey the contextualized meaning of each target token. The parameter weights in the language model are  optimized to minimize the negative log-likelihood of the ground-truth source token $t^0_{i}$ for each target token $t^*_{i}$:
\begin{align}
    \mathcal{J} =  \sum\limits_{i=1}^{N}-\log \frac{\text{sim}(\textbf{H}(t^0_{i}),\textbf{h}(t^*_{i}|c^*_i))}{\sum\limits_{j=1}^{N}\text{sim}(\textbf{H}(t^0_{j}),\textbf{h}(t^*_{i}|c^*_i))}\label{eq:wse-objective}
\end{align} 
Here $\text{sim}(\cdot, \cdot)$ can be either a prototype-based similarity function in Eq.\ref{eq:prototype-sim}, or its exemplar-based counterpart specified in Eq.\ref{eq:exemplar-sim}. 

\section{Data}
\subsection{Dataset of polysemous word usages}
We construct our WSE dataset by collecting naturalistic usage instances of English polysemous words from the Wikitext-103 linguistic corpus \cite{merity2016pointer} that is commonly used as a language modeling benchmark. We first extract the sentences and lemmatize the corpus using SpaCy. We then apply a state-of-the-art word disambiguation algorithm by \citet{bevilacqua2020breaking} on each sentence to annotate each of its token with one of its associated WordNet synset IDs as the sense label \cite{miller1995wordnet}. We construct a polysemous English word vocabulary by taking word lemma types that satisfy the following conditions: 1) the word type has least 2 different senses detected in the corpus; 2) each mention of the word type has one of the four part-of-speech categories as detected by SpaCy: noun, verb, adjective, or adverb; 3) each sense of the word type has at least 10 mentions in the corpus. This process yields a large repertoire of 7,599 polysemous word types with a total number of 1,470,211 usage sentences, and an average number of 4.27 senses per word type. 

\subsection{Partioning polysemous word types}
To construct and evaluate our WSE framework, we partition each polysemous word types into multiple source-target pseudo-token pairs. In particular, for each word type $w$ with $n$ senses, we randomly choose one sense as the target sense $s^*$, and the remaining $n-1$ senses as the source senses. A source-target token pair is then created, which replace $w$ in usage sentences based on their sense labels following the procedures described in Section~\ref{type-partition}. We repeat this partitioning process 5 times so that each word type with at least 5 senses will have 5 distinct senses chosen as target, and for words with less than 5 senses, the 5 target senses will be sampled with replacement from its sense inventory. Each partition will therefore create $2\times7,599=15,198$ pseudo-tokens.

\section{Evaluation and results}
\subsection{Experimental setup}
We use a transformer model with the same architecture as BERT-base-uncased \cite{devlin-etal-2019-bert} as the main language model
in our WSE framework. The parameter weights of our language models are randomly initialized to prevent any information smuggling (i.e., the models are trained from scratch). In the masked language modeling training stage on replaced usage sentences, we increase the vocabulary size of each model by replacing all polysemous word types in our WSE dataset vocabulary with their partitioned pseudo-tokens, and add rows to embedding layer and final classification layer of the BERT model accordingly. Five language models are trained independently, one for each set of partitioned tokens as described in section 4.2. During sense-extensional semantic space learning, we randomly choose 70\% of the original polysemous word types and take usage sentences containing their partitioned tokens as the training set. Sentences containing partitioned tokens spawned by the remaining 30\% word types will be taken as the test set, so that there is no overlap in the vocabulary of partitioned tokens or their parent word types between training and testing.\footnote{See Appendix \ref{appendix:wse-model-implementation} for more implementation details.} 
\subsection{Baseline models}
We also compare the performance of our WSE models against a set of baseline models  without chaining-based inference mechanisms: 1) a BERT-MLM baseline ignores the intended meaning information and predicts $P(t^0|\textbf{m}(t^*|c^*))$ as the infilling probability of $t^0$ under context $c^*$ with $t^*$ replaced by a masking placeholder; 2) a BERT-STS baseline computes the contextualized representation $\textbf{h}(t^0|c^*)$ of each candidate source token $t^0$ under $c^*$, and calculates $P(t^0|\textbf{m}(t^*|c^*))$ as proportional to the cosine similarity between $\textbf{h}(t^0|c^*)$ and the contextualized embedding $\textbf{h}(t^*|c^*)$ of the target token under the same context (i.e. based on the semantic textual similarity between contextualized meanings of $t^0$ and $t^*$). Both baselines are built on the same BERT encoder just as the two chaining-based WSE models. We also consider a random baseline that  randomly draws a source token from the set of alternative candidate tokens.

\begin{table*}[h!]
\centering
\begin{tabular}{@{}lcccc@{}}
\toprule
\multicolumn{1}{c}{\multirow{2}{*}{Model}} & \multicolumn{2}{c}{Mean reciprocal rank}                          & \multicolumn{2}{c}{Mean precision}                                \\ \cmidrule(l){2-5} 
\multicolumn{1}{c}{}                       & \multicolumn{1}{c}{Unsupervised} & \multicolumn{1}{c}{Supervised} & \multicolumn{1}{c}{Unsupervised} & \multicolumn{1}{c}{Supervised} \\ \midrule
Random Baseline                            & 5.21                             & 5.21                           & 1.00                             & 1.00                           \\
BERT-STS                                   & 11.89 (0.54)                     & 33.55 (0.97)                   & 14.02 (0.58)                      & 25.57 (0.79)                    \\
BERT-MLM                                   & 15.57 (0.60)                     & 37.09 (0.92)                   & 16.34 (0.70)                      & 28.99 (0.63)                   \\
WSE-Prototype                              & 29.96 (0.77)                     & 48.04 (1.03)                   & 21.50 (0.44)                      & 35.78 (1.16)                   \\
WSE-Exemplar                               & \textbf{34.25 (0.99)}            & \textbf{53.79 (1.07)}          & \textbf{29.17 (1.28)}             & \textbf{37.82 (1.45)}         
\end{tabular}
\caption{Summary of model mean precision and MRR-100 scores (\%) for word sense extension. Numbers after $\pm$ are standard deviations over 5 sets of independently partitioned source-target token pairs.}
\label{table:wse-results}
\end{table*}

\begin{table*}[h!]
\resizebox{\textwidth}{!}{%
\begin{tabular}{|lll|}
\hline
\multicolumn{1}{|l|}{Model} & \multicolumn{1}{l|}{Top-5 predicted words (source tokens)} & Predicted rank of ground-truth source token \\ \hline
\multicolumn{3}{|l|}{\begin{tabular}[c]{@{}l@{}}Word: \textit{cover}; target sense: \textbf{be responsible for reporting news} \\ Usage context: Generally, only reporters who \textit{cover} breaking news are eligible.\end{tabular}} \\ \hline
\multicolumn{1}{|l|}{BERT-MLM} & \multicolumn{1}{l|}{work, take, write, report, send} & \databar{red!40}{44}{54/100} \\ \hline
\multicolumn{1}{|l|}{WSE-Exemplar} & \multicolumn{1}{l|}{practice, report, supervise, cover, know} & \databar{blue!40}{180}{4/100} \\ \hline
\multicolumn{3}{|l|}{\begin{tabular}[c]{@{}l@{}}Word: \textit{cell}; target sense: \textbf{a room where a prisoner is kept}\\ Usage context: on the eve of his scheduled execution, he committed suicide in his \textit{cell} with a smuggled blasting cap ...\end{tabular}} \\ \hline
\multicolumn{1}{|l|}{BERT-MLM} & \multicolumn{1}{l|}{place, house, room, bedroom, hall} & \databar{red!40}{82}{63/100} \\ \hline
\multicolumn{1}{|l|}{WSE-Exemplar} & \multicolumn{1}{l|}{room, cell, bedroom, pocket, pyjamas} & \databar{blue!40}{196}{2/100}  \\ \hline
\multicolumn{3}{|l|}{\begin{tabular}[c]{@{}l@{}}Word: \textit{grasp}; target sense: \textbf{to get the meaning of}\\ Usage context: Madonna later acknowledged that she had not \textit{grasped} the concept of her mother dying.\end{tabular}} \\ \hline
\multicolumn{1}{|l|}{BERT-MLM} & \multicolumn{1}{l|}{understand, remember, enjoy, comprehend, keep} & \databar{red!40}{28}{82/100} \\ \hline
\multicolumn{1}{|l|}{WSE-Exemplar} & \multicolumn{1}{l|}{understand, resolve, know, get, convey} & \databar{blue!40}{124}{43/100} \\ \hline
\end{tabular}%
}
\caption{Example predictions made by the WSE-Exemplar model and the BERT-MLM baseline
(supervised version). The top-5 predicted source tokens are translated into the (lemmatized) parent words from which they are partitioned.
}
\label{table:wse-example-outputs}
\end{table*}

\subsection{Evaluation on WSE}
We first evaluate our models on the task of predicting source partitioned tokens formulated in Eq.\ref{eq:wse-inference}. At each trial, for each target token $t^*_w$ partitioned from $w$, we present the model with the ground-truth source token $t^0_w$ partitioned from the same word $w$, and 99 negative candidate source tokens $t^0_{w'}$ spawned from different polysemous word types $w'$. Both the ground-truth source token and the negative candidates are sampled from the evaluation set for sense-extensional semantic space learning. We assess each model in two settings: an unsupervised version of a model that does not learn from the training set of WSE, and a supervised version that is trained on the training set of sense extensional space learning. The BERT encoders of the supervised versions of two BERT baselines are trained using the same objective function and data as defined in Section~\ref{sense-extension-space}. 

We quantify model performance with two metrics: 1) the mean precision is the percentage of cases where a model correctly predicts the ground-truth source token as the most likely candidate, and 2) the mean reciprocal rank (MRR-100) is the averaged multiplicative inverse of the ranks of the ground-truth source tokens in all evaluation examples. Table \ref{table:wse-results} summarizes the overall results in the five sets of independently partitioned tokens. We make several observations: 1) all BERT-based models perform substantially better than chance even without explicit training on WSE. This can be explained by the fact that many polysemous word types in our dataset have very fine-grained WordNet senses, so that the target senses chosen from its sense inventory are often highly similar or even hardly distinguishable from the some source senses of the same word; 2) all BERT-based models benefit from learning a sense-extensional semantic space, suggesting the presence of regularity shared among examples of sense extension across word types; 3) both chaining-based WSE models consistently outperform other baselines in both the unsupervised and supervised settings. The exemplar-based WSE models generally outperform than their prototype-based counterparts, suggesting that word sense extension depends on the speaker's sensitivity to the semantic similarity between the intended meaning and the individual (exemplar) usages.    

\begin{figure}[h!]
 \includegraphics[clip,width=\columnwidth]{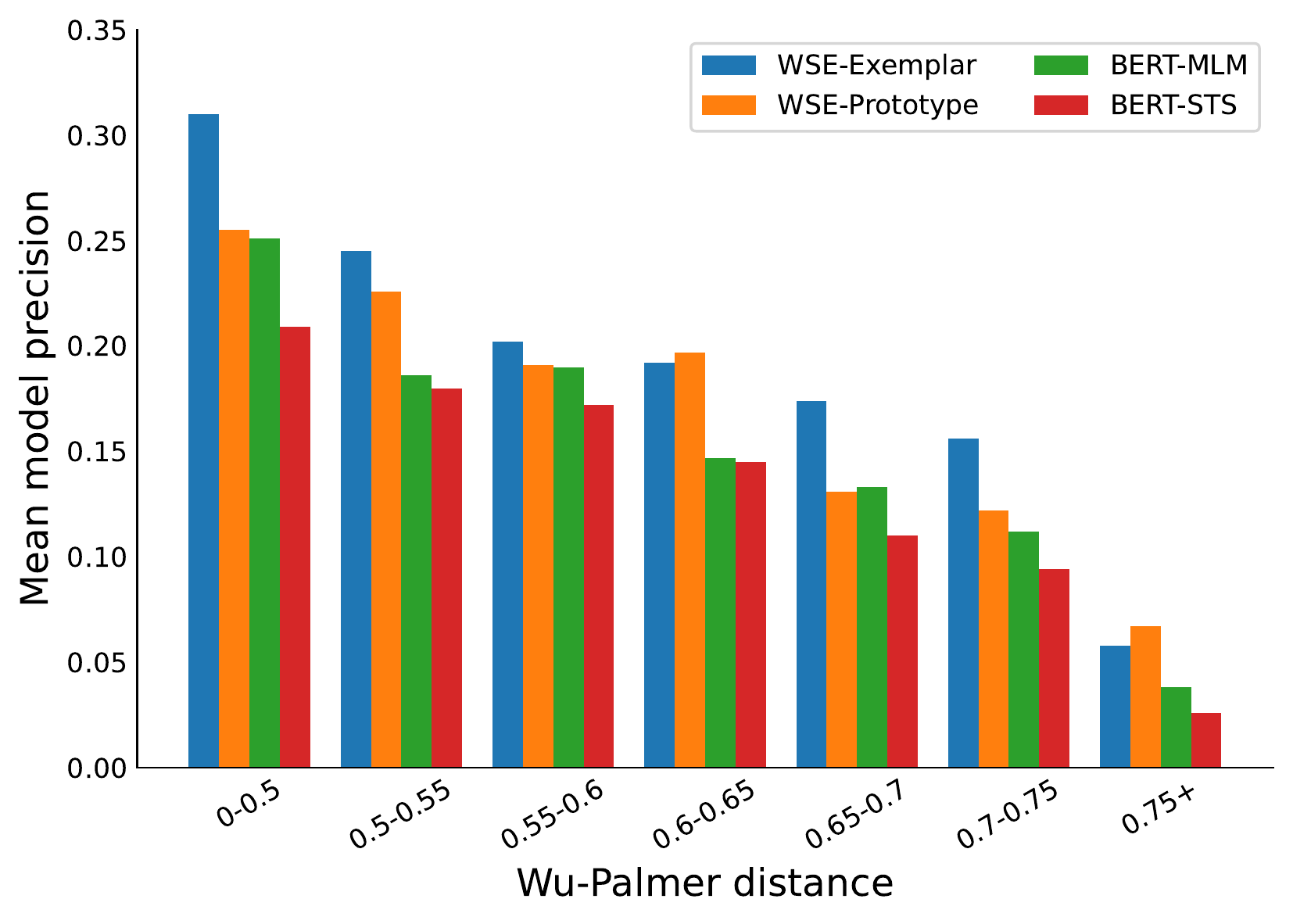}
\caption{Mean model precision vs. Wu-Palmer distance between WordNet synsets associated with fully-partitioned tokens.}\label{figure-concept-rel-vs-performance}
\end{figure}

Table \ref{table:wse-example-outputs} shows example predictions on sample polysemous words made by the supervised exemplar-based WSE model and the supervised BERT-MLM baseline. The WSE model successfully predicts many types of sense extension, such as metaphorical senses for both the verb \textit{cover} example and the noun \textit{cell}. In contrast, the BERT-MLM baseline shows a greater tendency to predict a literal paraphrase for a partitioned token. Still, both WSE and baseline models  struggle with predicting some usages that involve strong non-literal sense extension (e.g., the \textit{grasp} example).

\begin{table*}[h!]
\centering
\resizebox{0.9\textwidth}{!}{%
\begin{tabular}{@{}lcccccccccc@{}}
\toprule
                          & Dev           & \multicolumn{4}{c}{Test Datasets}                             & \multicolumn{5}{c}{Concatenation of Test Datasets}                            \\ \midrule
                          & SE07          & SE02          & SE03          & SE13          & SE15          & Nouns         & Verbs         & Adj.          & Adv.          & ALL           \\ \midrule
WordNet S1                & 55.2          & 66.8          & 66.2          & 63.0          & 67.8          & 67.6          & 50.3          & 74.3          & 80.9          & 65.2          \\
Most frequent sense (MFS) & 54.5          & 65.6          & 66.0          & 63.8          & 67.1          & 67.7          & 49.8          & 73.1          & 80.5          & 65.5          \\ \midrule
BERT-linear               & 68.6          & 75.2          & 74.7          & 70.6          & 75.2          & 74.6          & 63.6          & 78.6          & 87.0          & 73.5          \\
+ WSE-Prototype           & \textbf{70.9} & \textbf{78.0} & \textbf{75.2} & \textbf{71.2} & \textbf{77.9} & \textbf{75.5} & \textbf{66.1} & 78.9          & \textbf{87.1} & \textbf{76.4} \\
+ WSE-Exemplar            & 70.5          & \textbf{78.0} & 75.1          & \textbf{71.2} & 77.7          & 74.8          & 65.8          & \textbf{79.2} & 86.4          & 75.3          \\ \midrule
BEM                       & 74.3          & 78.8          & \textbf{77.4} & 79.6          & 80.9          & 81.5          & 68.5          & 82.8          & \textbf{87.1} & 78.8          \\
+ WSE-Prototype           & \textbf{74.9} & \textbf{80.2} & 75.9          & \textbf{81.2} & 81.1          & \textbf{82.5} & \textbf{70.2} & \textbf{83.9} & \textbf{87.1} & \textbf{80.1} \\
+ WSE-Exemplar            & 74.5          & 80.0          & 76.1          & \textbf{81.2} & \textbf{81.7} & 81.4          & 69.1          & 81.2          & 86.4          & 79.2          \\ \bottomrule
\end{tabular}
}
\caption{F1-scores (\%) for fine-grained all-words WSD task on the evaluation framework by \cite{raganato2017word}.}
\label{wsd-main-results}
\end{table*}

\begin{table*}[h!]
\centering
\resizebox{0.9\textwidth}{!}{%
\begin{tabular}{@{}lll@{}}
\toprule
WSD test example                                                                                                                                                                                                                                           & BEM prediction (no WSE)                                                                                                & BEM prediction (with WSE)                                                                                               \\ \midrule
\begin{tabular}[c]{@{}l@{}}\textbf{Context}: The purpose of education \\ is to encourage young men and women \\ to realize their \textbf{\textit{full}} academic potential.\\ \textbf{Target sense training frequency}: 0\end{tabular}             & \begin{tabular}[c]{@{}l@{}} containing as much or \\ as many  as is possible (\xmark)\end{tabular} & \begin{tabular}[c]{@{}l@{}}complete in extent or degree (\cmark)\end{tabular} \\ \midrule
\begin{tabular}[c]{@{}l@{}}\textbf{Context}: Haney felt like \textit{\textbf{shrinking}} out of sight, \\ but he was already trapped in the \\ corner with the wiry, dark little man.\\ \textbf{Target sense training frequency}: 1\end{tabular} & \begin{tabular}[c]{@{}l@{}}reduce in size/physically (\xmark)\end{tabular} & \begin{tabular}[c]{@{}l@{}}draw back with fear or pain (\cmark)\end{tabular} \\ \bottomrule
\end{tabular}
}
\caption{Examples of context and definitions of WSD-model predicted senses. The bold italic words in context are disambiguated by
the BEM model before and after training on WSE.}
\label{table:wsd-output-examples}
\end{table*}

\subsection{Sense relatedness and model predictability}
Prior work in psycholinguistics suggests that both adults and children often find it easy to infer a new intended meaning of a word if they can access a highly related conventional sense of that word to constrain their interpretation \cite{clark1983understanding,klepousniotou2008making,rodd2012learning}. We examine whether our WSE models exhibit human-like sensitivity to the conceptual relatedness between existing and novel word senses. For each source-target partitioned token pair $(t^0, t^*)$, we quantify their degree of conceptual relatedness as the mean Wu-Palmer semantic distance \cite{wu1994verbs} between the WordNet synset of the target sense denoted by $t^*$ and the synset of each existing source sense of $t^0$. Figure \ref{figure-concept-rel-vs-performance} shows the performance of 4 WSE model variants on predicting sense pairs binned with respect to their degree of conceptual similarity. We observe that the WSE models generally make better predictions on source-target token pairs that are semantically more related (e.g., metonymy), and perform less well on examples where the target sense is conceptually very different to the existing source senses (e.g., strong metaphor or homonymy).

\subsection{Application of WSE to WSD}

As a final step, we show that state-of-the-art word sense disambiguation models can benefit from the word sense extension framework. We evaluate WSD models on the standard WSD evaluation framework proposed by \cite{raganato2017word}, where in each trial, the model is given an input sentence and is asked to assign WordNet sense labels for a subset of tokens within the sentence. We consider two BERT-based WSD models: 1) a BERT-linear model that learns a linear classifier for WSD on top of a frozen BERT encoder. This model does not incorporate gloss information, and cannot predict novel senses that do not appear in training;  2) a bi-encoder model (BEM) by \cite{blevins-zettlemoyer-2020-moving} independently encodes input sentences with target words and sense glosses via two encoders, each of which are initialized with BERT-base. The contextualized embedding of the target word then takes dot product with the gloss embedding of each candidate sense, and the model predicts the sense with highest dot product score with the embedded target word. This model has been shown to yield impressive results on WSD examples with rare senses. 

To integrate WSE into WSD, we  fine-tune the BERT encoder of each WSD model on the WSE training set of Wikitext-103 usage sentences via the objective in Eq. \ref{eq:wse-objective}, which can be formulated as either a prototype model or an exemplar model. Unlike the case of WSE evaluation, here we use pretrained BERT-base-uncased encoders and keep the original word form of each polysemous word without partitioning it into source-target token pairs. The resulting BERT encoder is then taken to learn one of the two WSD models described above, and evaluated on  WSD tasks. For BEM, both encoders are initialized as the BERT-base fine-tuned on WSE. Since the sense labels of usage sentences in the WSE dataset are not fed to BERT during training, none of the models has access to any  usage examples of target senses in the WSD test set.

\begin{table}[h!]
\centering
\resizebox{0.9\columnwidth}{!}{%
\begin{tabular}{@{}lccc@{}}
\toprule
                     & \multicolumn{3}{c}{Sense frequency} \\
\multicolumn{1}{c}{} & High    & Few-shot    & Zero-shot   \\ \midrule
BERT-linear          & 81.7    & 54.4        & 53.6        \\
+ WSE                & 82.3    & 60.1        & 53.6        \\ \midrule
BEM                  & 86.8    & 77.7        & 67.8        \\
+ WSE                & 86.6    & 79.6        & 71.5        \\ \bottomrule
\end{tabular}}
\caption{F1-score (\%) on subsets of the WSD test dataset grouped by target sense frequency in  SemCor corpus.}
\label{table:wsd-f1-by-freq}
\end{table}

Table \ref{wsd-main-results} reports overall results on the WSD datasets under the standard F1-score. We also include the performance of two simple baselines: 1) WordNet S1 always predicts the first sense, and 2) MFS always predicts the most frequent sense in the training data. We found that chaining-based WSE models improve the performance of the two BERT-based WSD models on almost every test subset, as well as on all POS categories except for the adverb class. These results show that WSE may serve as useful pretraining  for improving WSD models both with and without access to gloss information.  

{\bf Rare word-sense pairs.} We hypothesize that WSE improves WSD because learning word sense extension helps the model to better interpret rare senses that bear systematic semantic relations with more conventional senses. Table \ref{table:wsd-f1-by-freq} shows the performance of WSD models grouped by the frequency of the target word sense in the WSD training set. We define zero-shot test cases as target senses that never appear during WSD training, and few-shot test cases as those with 1 to 10 mentions, and high-frequency senses as those with more than 10 training mentions. The BERT-linear model resort to a most frequent sense heuristic for zero-shot examples, since it cannot learn a classification layer embedding for previously unattested senses. We observe that all WSD models trained on WSE yield substantially greater improvement for few-shot and zero-shot test cases, while maintaining high performance on the more frequent cases. Table \ref{table:wsd-output-examples} shows test examples where incorrect predictions of BEM are improved with WSE integration. These examples often exhibit regular semantic relations between target and conventional senses of a word (e.g., the relation between physical size and amount that underlies the two attested senses of \textit{full}).

\section{Conclusion}
We have presented a  framework for word sense extension that supports lexical items to extend to new senses in novel context. Our results show that chaining provides a general mechanism for automated novel sense extension in natural context, and learning a transformed sense-extensional space enables systematic generalization to a certain degree. We also show that word sense extension improves the performance of transformer-based WSD models particularly on rare word senses. Future work may extend our framework in several ways, such as how to better model systematic word sense extension, and do so over time and in different languages.

\section{Ethical considerations}
We discuss the limitations and potential risks of our work.
\subsection{Limitations}
Our current framework does not explicitly consider the temporal order via which word senses have emerged. In particular, in the data collection step, we construct source-target token pairs for each word type by randomly sampling a target sense from its sense inventory. An alternative and more realistic approach would be to sort all senses of a word chronologically by their times of emergence in history, and use the model to incrementally predict each sense of a word based on usages of its older senses. However, we found that it is infeasible to find accurate timestamps of senses in natural corpora at a comprehensive scale. Another approach is to have human annotators evaluate the plausibility of each ground-truth source-target token pairs that are automatically created in our data collection pipeline, which is a potential area for future consideration.

\subsection{Potential risks}
 All scientific artifacts in this study have been made publicly available and are consistent with their intended use and access conditions. We acknowledge that our focus on English might introduce linguistically or culturally specific biases in model-generated outputs. For instance, we observe that the WSE models trained on English sentences learn to generate a metaphorical expression ``to \textit{spend} some time'' for the English verb \textit{spend}, which is common in English  but differ in other languages (e.g., Hungarian speakers instead tend to say ``to \textit{fill} some time'' as in \citealt{kovecses2010metaphor}). We believe that by training WSE models cross-linguistically to cover various innovative lexical uses should help mitigate this issue.

\section{Acknowledgements}

This work was supported by a NSERC Discovery Grant RGPIN-2018-05872.

\bibliography{anthology,custom}
\bibliographystyle{acl_natbib}

\appendix



\section{Implementations of WSE models}
\label{appendix:wse-model-implementation}
We use the BERT-base-uncased configuration provided by Hugging Face \cite{wolf2020transformers} to initialize all BERT-based WSE models (two baselines and two chaining-based models). During MLM pretraining of BERT models on replaced usage sentences by partitioned pseudo-tokens, we randomly mask 15\% of tokens in each sentence, and train each model on predicting the masked tokens. We add all partitioned pseudo-tokens as special tokens into the vocabulary of the BERT tokenizer, so each pseudo-token will be encoded as a whole in the input sequence. Learning is performed using the Adam optimizer \cite{kingma2015adam}, with a learning rate of 5e-5 and a batch size of 128, for 8 epochs (after which all models achieved highest evaluation accuracy). During sense-extensional semantic space learning, both exemplar-based and prototype-based models are trained on the objective function in Eq.\ref{eq:wse-objective} using Adam, with a mini-batch size of 16 and a learning rate of 2e-5, for 8 epochs (after which all models achieved highest evaluation accuracy). All experiments are run on machines with 4 NVIDIA Tesla V100 GPUs, with an average training time of 30 minutes per epoch for MLM pretraining, and 12 minutes per epoch for sense-extensional semantic space learning.

\end{document}